\documentclass{book}    
\usepackage{piers}  
\usepackage[]{graphicx}
\graphicspath{{Figures/}}
\usepackage{subfigure}
\usepackage{amsmath}
\usepackage{amssymb}
\usepackage{epsfig}
\usepackage{cite}
\usepackage{color}
\usepackage{balance}
\usepackage{pgfplots}
\usepackage{wrapfig}
\usepackage{lipsum}
\usepackage{hyperref}

\begin{document}

\title{Correlating Satellite Cloud Cover with Sky Cameras}
\maketitle

\begin{authors}

{\bf Shilpa Manandhar}$^{1}$ $^{*}$, {\bf Soumyabrata Dev}$^{2}$, {\bf Yee Hui Lee}$^{3}$, {\bf and Yu Song Meng}$^{4}$\\
\medskip

$^{1}$Nanyang Technological University Singapore, Singapore 639798, email: \texttt{shilpa005@e.ntu.edu.sg}\\

$^{2}$Nanyang Technological University Singapore, Singapore 639798, email: \texttt{soumyabr001@e.ntu.edu.sg}\\

$^{3}$ Nanyang Technological University Singapore, Singapore 639798, email: \texttt{EYHLee@ntu.edu.sg}\\

$^{4}$National Metrology Centre, Agency for Science, Technology and Research (A$^{*}$STAR), Singapore 118221, email: \texttt{meng\_yusong@nmc.a-star.edu.sg}

$^{*}$ Presenting author and Corresponding author

\end{authors}

\begin{paper}

\begin{piersabstract}
The role of clouds is manifold in understanding the various events in the atmosphere, and also in studying the radiative balance of the earth. The conventional manner of such cloud analysis is performed mainly via satellite images. However, because of its low temporal- and spatial- resolutions, ground-based sky cameras are now getting popular. In this paper, we study the relation between the cloud cover obtained from MODIS images, with the coverage obtained from ground-based sky cameras. This will help us to better understand cloud formation in the atmosphere -- both from satellite images and ground-based observations.

\end{piersabstract}

\psection{Introduction}

The Moderate Resolution Imaging Spectroradiometers (MODIS) installed on National Aeronautics and Space Administration (NASA) Earth Observing System’s (EOS) Terra and Aqua satellites is an excellent source of such important and long-term record of climate data. MODIS data are increasingly being used in weather related studies like, the PWV products available from MODIS are now used in severe weather simulations \cite{MODIS}, analyzing cloud optical properties~\cite{PIERS17b}, and study of aerosol properties \cite{aerosol}.
Amongst other useful products, one of the important contributions from MODIS images is \emph{cloud mask} that indicates the presence of cloud over a particular area. However, its use is limited because it can provide a top-view of cloud formation, effectively neglecting the low-lying clouds. It also has low temporal and spatial resolutions, as its satellites passes over Singapore only twice a day. Therefore, images captured using ground-based sky cameras are now slowly gaining popularity amongst the remote sensing analysts. We analyze and compare the cloud cover obtained from both satellite- and ground-based- images.

\psection{Data Collection}
\psubsection{Ground-based Cameras}

We designed and deployed our custom-built sky cameras at the rooftop of our university building ($1.3483^\circ$ N, $103.6831^\circ$ E). We refer our sky cameras as WAHRSIS, that stands for Wide Angled High Resolution Sky Imaging System~\cite{WAHRSIS,IGARSS2015}. Such ground-based sky camera captures the sky scene at an interval of $2$ minutes. It captures images in the visible-light spectrum, and have a higher temporal and spatial resolutions, as compared to the conventional satellite images. We use the ratio of \emph{red} and \emph{blue} color channels to detect clouds in the captured sky/cloud image~\cite{ICIP1_2014}. This captured sky/cloud images assist us in computing the cloud coverage, which is defined as the ratio of the sky scene covered by clouds.

\psubsection{MODIS images}
There are different levels of MODIS products available online ~\footnote{The MODIS data are available online at \url{https://ladsweb.modaps.eosdis.nasa.gov/archive/allData/?process=ftpAsHttp&path=allData}.}. For this paper we use MODIS level 5 products, which gives information on the cloud mask values. The cloud mask values calculated from MODIS products have spatial resolution of $1$ km{$^2$} ($1$ pixel) and is available twice a day ($4$ UTC and $7$ UTC). The value of cloud mask can take any values 0, 64, 128 or 192; 0 indicates 100 \% cloudy condition and 192 indicates the cloud free condition. For this paper, we take the average cloud coverage a certain pixel area and normalize it to the range of 0 and 100, such that the 0 represent no cloud and 100 represent full cloud conditions.

\psection{Experiments \& Results}
\psubsection{Comparison between WAHRSIS images and MODIS images}

\begin{figure}[htb]
\centering
\includegraphics[width=1\textwidth]{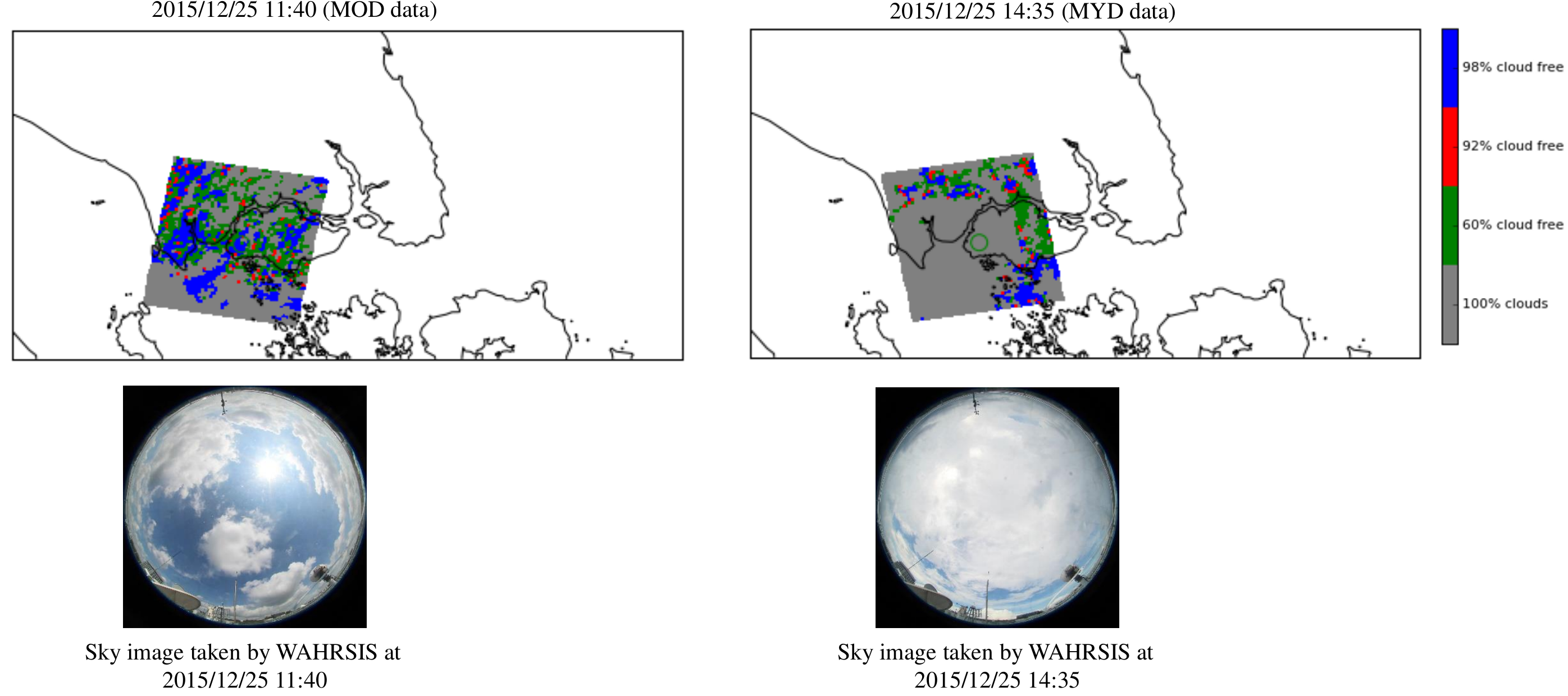}
\caption{Comparison between ground based sky images and satellite images taken by WAHRSIS at two different time stamps of 2015/12/25.}
\label{fig:Comp}
\end{figure}

Figure \ref{fig:Comp} shows the comparison between MODIS images and WAHRSIS images. For the comparison purpose, MOD and MYD MODIS images captured on 25-Dec-2012 are used. Here, four different colors are used to represent the cloud mask values; blue represents cloud mask values of 192 indicating 98 \% cloud free condition, green represents cloud mask values of 128 indicating 92 \% cloud free condition, red represents cloud mask values of 64 indicating 60 \% cloud free condition and grey represent cloud mask values of 0 indicating 100 \% cloudy conditions. A green circle, clearly visible on MODIS image taken on 2015/12/25 is used to represent the location of NTU (1.34$^{\circ}$, 103.68$^{\circ}$). The MODIS plot shows the sky condition taking NTU position as a center pixel.

We observe that the MODIS MOD image captured on 2015/12/25 at 11:40 AM shows that the sky is almost cloud free during that time, as the color plot shows majority of blue color. A WAHRSIS image taken at the same time is shown which also shows a clear sky condition. Similarly, for the same day, the MYD image captured at 14:35 PM shows a cloudy condition at NTU location as the area is represented by grey color in majority. And the sky image captured by WAHRSIS at the same time also shows fully cloudy condition. These preliminary results suggests a visual correlation between the cloud condition indicated by the MODIS images and that by the sky images. In the next section of the paper, we discuss the correlation in further details. 

\psubsection{Cloud coverage from sky images and cloud mask from MODIS}
In this experiment~\footnote{The source codes of these experiments are available online at \url{https://github.com/Soumyabrata/MODIS-cloud-mask}.}, all the MODIS data in the year $2015$ was considered. We process cloud mask values for 3 km $\times$ 3 km area (9 pixels) with the sky camera as the center location.  We compute the average of the entire $9$ pixel block (excluding invalid data points). We also compute the cloud coverage of the nearest sky camera image, captured by our sky camera. 

Figure~\ref{fig:CM-fig} shows the statistical analysis of the average cloud mask value with respect to the cloud coverage from images. We bin the average cloud mask values into $4$ distinct bins, as the actual cloud mask has $4$ distinct levels. 
 
\begin{figure}[htb]
\centering
\includegraphics[height=0.55\textwidth]{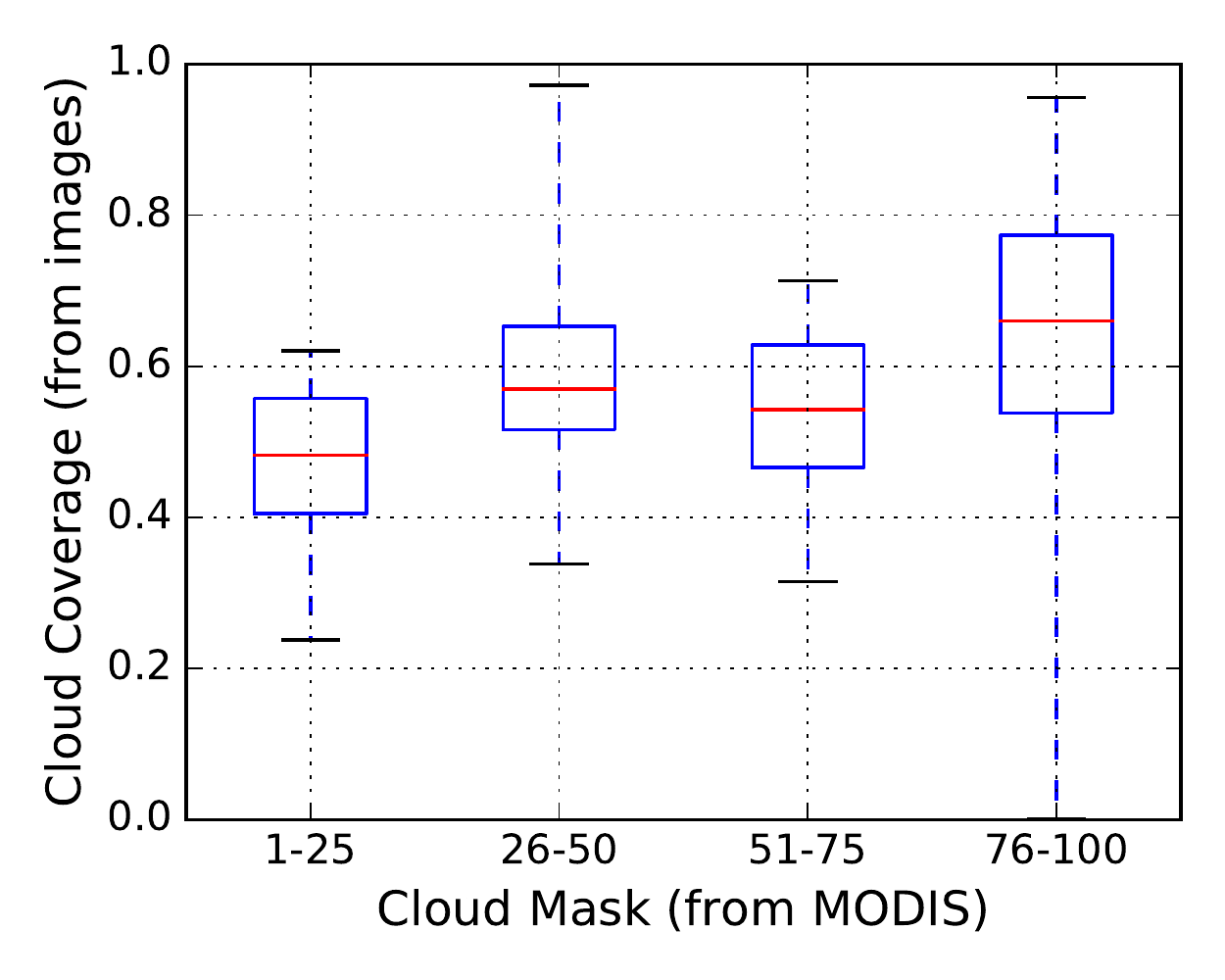}
\vspace{-0.5cm}
\caption{Statistical analysis of cloud mask (obtained from MODIS) with respect to cloud coverage (obtained from ground-based sky cameras).}
\label{fig:CM-fig}
\end{figure}

We observe the general trend between cloud mask and cloud coverage -- the cloud coverage increases with an increase in the cloud mask values. However, there is a higher variation in cloud coverage for higher cloud mask values. This is because of the possible area mismatch between MODIS and ground-based sky cameras.

\psection{Conclusion}
In this paper, cloud mask values obtained from MODIS images are compared to the cloud coverage calculated from sky images. A good agreement is found between the two which could be seen from the time series plot and the statistical plot as well. This analysis between satellite- and ground-based observations will provide the remote sensing analysts further insights into cloud formation, and understanding the role of clouds in the radiative balance of the earth. In the future, we plan to deploy multiple sky cameras employing advanced compression schemes~\cite{Deepu2015}, for continuous weather analysis.

\ack
The authors would like to thank Joseph Lemaitre for automatizing the acquisition and processing of MODIS multi bands images into a user-friendly framework.

\end{paper}


\begin{thebibliography}{1}

\bibitem{MODIS}
S.~H. Chen, Z.~Zhao, J.~S. Haase, and F.~Vandenberghe A~Chen~and,
\newblock ``A study of the characteristics and assimilation of retrieved
  {MODIS} total precipitable water data in severe weather simulations,''
\newblock {\em American Meteorological Society}, vol. 136, pp. 3608--–3628,
  Sept 2008.

\bibitem{PIERS17b}
S.~Manandhar, S.~Dev, Y.~H. Lee, and Y.~S. Meng,
\newblock ``Analyzing cloud optical properties using sky cameras,''
\newblock in {\em Proc. Progress In Electromagnetics Research Symposium
  (PIERS)}, 2017.

\bibitem{aerosol}
J.~Wei and L.~Sun,
\newblock ``Comparison and evaluation of different modis aerosol optical depth
  products over the beijing-tianjin-hebei region in china,''
\newblock {\em Journal of Selected Topics in Applied Earth Observations and
  Remote Sensing}, vol. 10, no. 3, pp. 835--844, Mar 2017.

\bibitem{WAHRSIS}
S.~Dev, F.~M. Savoy, Y.~H. Lee, and S.~Winkler,
\newblock ``{WAHRSIS}: A low-cost, high-resolution whole sky imager with
  near-infrared capabilities,''
\newblock in {\em Proc. IS\&T/SPIE Infrared Imaging Systems}, 2014.

\bibitem{IGARSS2015}
S.~Dev, F.~M. Savoy, Y.~H. Lee, and S.~Winkler,
\newblock ``Design of low-cost, compact and weather-proof whole sky imagers for
  {High-Dynamic-Range} captures,''
\newblock in {\em Proc. International Geoscience and Remote Sensing Symposium
  (IGARSS)}, 2015, pp. 5359--5362.

\bibitem{ICIP1_2014}
S.~Dev, Y.~H. Lee, and S.~Winkler,
\newblock ``Systematic study of color spaces and components for the
  segmentation of sky/cloud images,''
\newblock in {\em Proc. International Conference on Image Processing (ICIP)},
  2014, pp. 5102--5106.

\bibitem{Deepu2015}
C.~J. Deepu and Y.~Lian,
\newblock ``A joint {QRS} detection and data compression scheme for wearable
  sensors,''
\newblock {\em IEEE Transactions on Biomedical Engineering}, vol. 62, no. 1,
  pp. 165--175, Jan. 2015.

\end{thebibliography}
\end{document}